\documentclass{article}
\usepackage{ijcai16}
\usepackage{times}
\usepackage{latexsym}
\usepackage{pgfplots}
\usepgfplotslibrary{external}
\usepackage{multirow}
\usepackage{amsmath}
\usepackage{graphicx}
\usepackage{subfig}
\usetikzlibrary{patterns}
\usetikzlibrary{pgfplots.colormaps}
\allowdisplaybreaks
\DeclareMathOperator*{\argmin}{arg\,min}

\title{Bridging LSTM Architecture and the Neural Dynamics during Reading}

\author{
Peng Qian \quad Xipeng Qiu\thanks{Corresponding author.} \quad Xuanjing Huang\\
Shanghai Key Laboratory of Intelligent Information Processing, Fudan University\\
School of Computer Science, Fudan University\\
825 Zhangheng Road, Shanghai, China\\
\{pqian11, xpqiu, xjhuang\}@fudan.edu.cn
}

\begin{document}

\maketitle

\begin{abstract}
Recently, the long short-term memory neural network (LSTM)  has attracted wide interest due to its success in many tasks. LSTM architecture consists of a memory cell and three gates, which looks similar to the neuronal networks in the brain. However, there still lacks the evidence of the cognitive plausibility of LSTM architecture as well as its working mechanism.
In this paper, we study the cognitive plausibility of LSTM by aligning its internal architecture with the brain activity observed via fMRI when the subjects read a story. Experiment results show that the artificial memory vector in LSTM can accurately predict the observed sequential brain activities, indicating the correlation between LSTM architecture and the cognitive process of story reading.
\end{abstract}

\section{Introduction}

In recent years, biologically-inspired artificial neural networks have become a focused topic in the field of computer science \cite{hinton2006reducing,bengio2009learning,schmidhuber2015deep}. Among the various network architectures, long short-term memory neural network (LSTM) \cite{hochreiter1997long} has attracted recent interest and gives state-of-the-art results in many tasks, such as time series prediction, adaptive robotics and control, connected handwriting recognition, image classification, speech recognition, machine translation, and other sequence learning problems \cite{schmidhuber2015deep}. LSTM is an extension of the simple recurrent neural network (RNN). It employs three gate vectors to filter information and a memory vector to store the history information. This mechanism can help encode long-term information better than simple RNN. Despite the biological inspiration of the architecture desgin of LSTM \cite{o2006making} and some efforts in understanding LSTM memory cell \cite{karpathy2015visualizing}, there still lacks the evidence of the cognitive plausibility of LSTM architecture as well as its working mechanism.

In this paper, we relate LSTM struture with the brain activities during the process of reading a story. In parallel with the fMRI experiment \cite{wehbe2014simultaneously}, we train a LSTM neural network and use it to generate the sequential representation of the same story. By looking for the potential alignment between the representations produced by LSTM and the neural activities recorded by fMRI at the same time, we are able to explore the cognitive plausibility of LSTM architecture {\it per se}.

Although some previous works \cite{mitchell2008predicting,pereira2011generating,pereira2013using,schwartz2013mapping,devereux2010using,murphy2012selecting} have tried to use computational models to decode the human brain activity associated with the meaning of words, most of them focused on the isolate words. Recently, \cite{wehbe2014aligning} studied the alignment between the latent vectors used by neural networks and brain activity observed via Magnetoencephalography (MEG) when subjects read a story. Their work just focused on the alignment between the word-by-word vectors produced by the neural networks and the word-by-word neural activity recorded by MEG.

Our main contributions can be summarized as follows. First, we show that it might be possible to use brain data to understand, interpret, and illustrate what is being encoded in the LSTM architecture, by drawing parallels between the model components and the brain processes; Second, we perform an empirical study on the gating mechanisms and demonstrate the superior power of the gates except the forget gates.

\section{Long Short-Term Memory Neural Network}


A recurrent neural network (RNN) \cite{Elman:1990} is able to process a sequence of arbitrary length by recursively applying a
transition function to its \emph{internal hidden state vector} $h_t$ of the input sequence. The activation
 of the hidden state $h_t$ at time-step $t$ is computed as a function $f$ of the current input symbol
  $x_t$ and the previous hidden state $h_{t-1}$
\begin{equation}
    h_t=
   \begin{cases}
   0 &\mbox{$t=0$}\\
   f(h_{t-1},x_t) &\mbox{otherwise}
   \end{cases}
\end{equation}

In a classic recurrent neural network, the gradient may blow up or decay exponentially over the time. Therefore, LSTM was proposed in \cite{hochreiter1997long} as a solution to the vanishing gradient problem. The basic unit of LSTM consists of three gates and a memory cell, which is designed in analogy to the psychological foundation of the human memory. A number of minor modifications to the standard LSTM unit have been made. While there are numerous LSTM variants, here we describe the implementation used by \cite{graves2013generating}.



LSTM unit has a memory cell and three gates: input gate, output gate and forget gate.
Intuitively, at time step $t$, the input gate $i_t$ controls how much each unit is updated, the output gate $o_t$ controls the exposure of the internal memory state, and the forget gate $f_t$ controls the amount of which each unit of the memory cell is erased. The memory cell $c_t$ keeps the useful history information which will be used for the next process.

Mathematically, the states of LSTM are updated as follows:
\begin{align}
& i_t = \sigma(W_{xi}x_t+W_{hi}h_{t-1}+W_{ci}c_{t-1}+b_i),\\
& f_t = \sigma(W_{xf}x_t+W_{hf}h_{t-1}+W_{cf}c_{t-1}+b_f),\\
& c_t = f_{t} \odot c_{t-1}+i_{t} \odot \tanh({W_{xc}x_t+W_{hc}h_{t-1}}+b_c),\\
& o_t = \sigma(W_{xo}x_t+W_{ho}h_{t-1}+W_{co}c_{t-1}+b_o),\\
& h_t = o_t \odot \tanh(c_t),
\end{align}
where $x_t$ is the input vector at the current time step, $\sigma$ denotes the logistic sigmoid function and $\odot$ denotes elementwise multiplication. Note that $W_{ci}$, $W_{cf}$ and $W_{co}$ are diagonal matrices.

When we use LSTM to model the linguistic input, such as sentences and documents, the first step is to represent the symbolic data into distributed vectors, also called embeddings \cite{bengio2003neural,collobert2008unified}.
Formally, we use a lookup table to map each word as a real-valued vector. All the unseen words are regarded by a special symbol and further mapped to the same vector.

\section{Methodology}

\begin{figure}[!t]
  \centering
  \includegraphics[width = 0.99\linewidth]{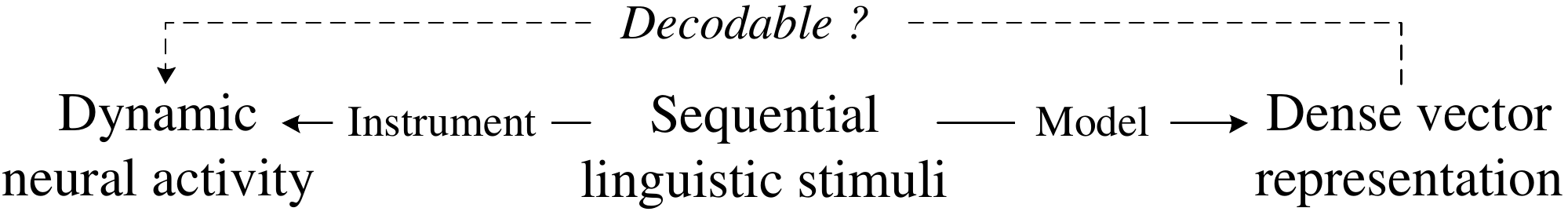}
  \caption{{The paradigm of brain-LSTM mapping experiment.}}\label{paradigm}
\end{figure}

Due to the complexity of LSTM, it is not always clear how to assess and compare its performances as they might be useful for one task and not the other. It is also not easy to interpret its dense distributed representations.

\begin{figure}[t]
  \centering
  \includegraphics[width = 0.99\linewidth]{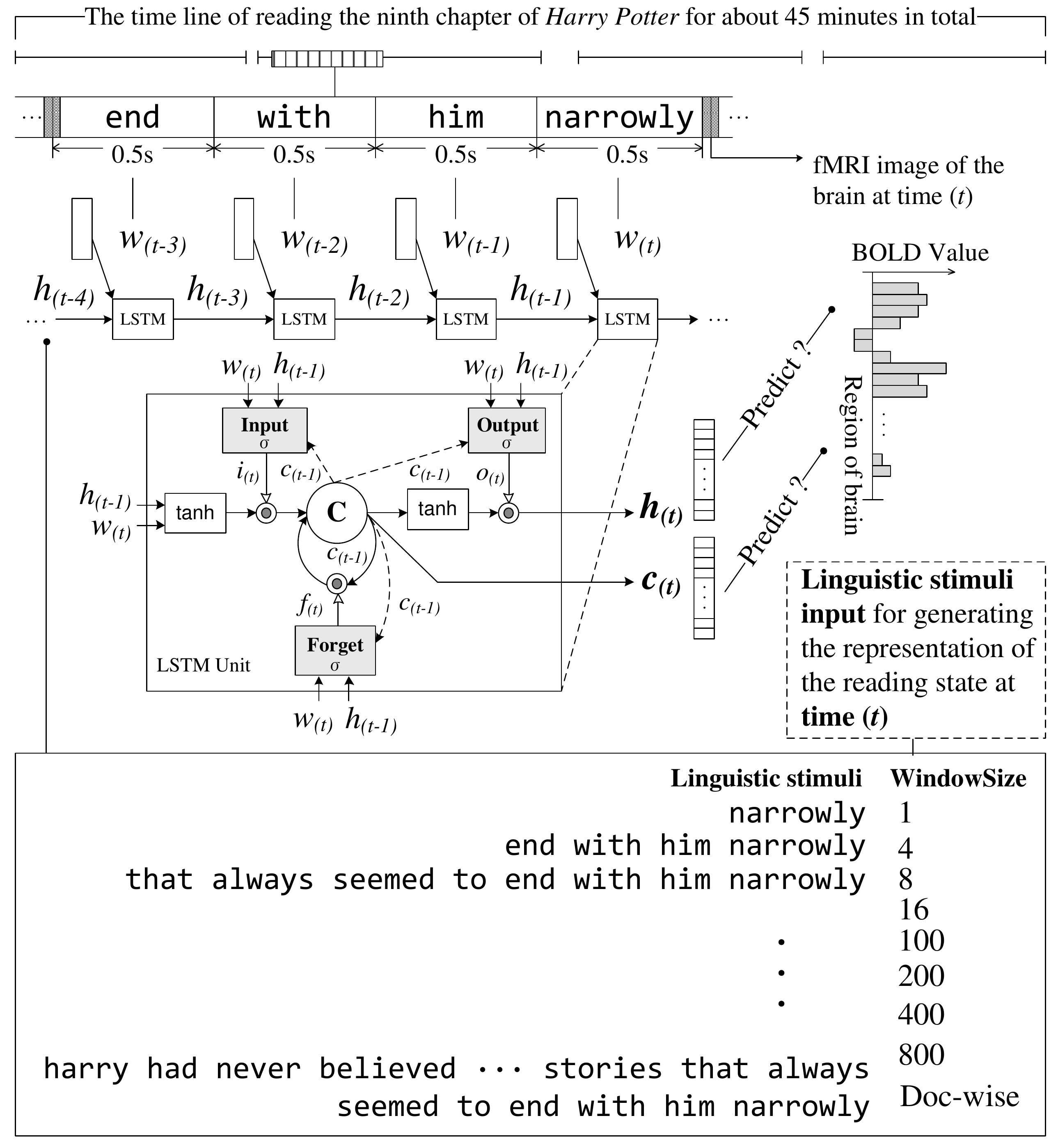}
  \caption{{An explanation of the analogy and alignment between LSTM mechanism and the process of reading a story chapter.}}\label{reading}
\end{figure}

In order to explore the correlation between the LSTM architecture and human cognitive process, we employ the paradigm of mapping the artificial representation of the linguistic stimuli with the real observed neural activity, as is explained in Figure \ref{paradigm}. One one hand, stimulated by a series of linguistic input, the neural response can be measured by the brain imaging techniques (e.g. EEG, fMRI, etc.). On the other hand, given the same series of the linguistic stimuli as the input information, an artificial model (e.g. recurrent neural network) also generates an abstract, continuous vector representation in correspondence with the real-time brain state. What would be attractive to us is whether there exists any linear mapping relationship between the model-based representation and the brain activity. This would guide us to a new direction of depicting the mechanism of model, specifically LSTM architecture in this paper. Figure \ref{reading} illustrates our experimental design.

%
%

In this section, we first briefly introduce the brain imaging data, and then we describe our experiment design.

\subsection{Brain Imaging Data}

The brain imaging data is originally acquired in \cite{wehbe2014simultaneously}, which recorded the brain activities of 8 subjects when they read the ninth chapter from the famous novel, {\it {Harry Porter and the Philosopher's Stone}} \cite{rowling1997harry}. The chapter had been segmented to words so that they can be presented to the subject at the center of the screen one by one, staying for 0.5 seconds each. Since the chapter is quite long and complicated, the whole chapter was divided into four sections. Subjects had short breaks between the presentation of the different sections. Each section started with a fixation period of about 20 seconds, during which the subjects stared at a cross in the middle of the screen. The total length of the four sections was about 45 minutes. About 5180 words were presented to each subject in the story reading task.

The brain activity data is collected by the functional Magnetic Resonance Imaging (fMRI), a popular brain imaging technique used in the cognitive neuroscience research. As fMRI displays poor temporal resolution, the brain activity is acquired every 2 seconds, namely every 4 words. Details of the data acquisition can be referred to \cite{wehbe2014simultaneously}. Of course, there are two potential limitations with fMRI. One is the low temporal resolution, compared with EEG. The other is that fMRI BOLD (Blood Oxygenation Level Dependent) signal is an indirect measurement of the neural activities. However, its high spatial resolution and non-invasive characteristic have made it a successful tool in cognitive neuroscience, especially for human subjects. Thus, we think that it is appropriate to measure the neural dynamics with fMRI, since it has been widely accepted by the academic community.

In Figure \ref{reading}, we summarize the basic information about the experiment setting of the story reading task. Taking one time step $t$ of the whole story time line as an example, the previous 4 words `end' ($w_{(t-3)}$), `with' ($w_{(t-2)}$), `him' ($w_{(t-1)}$), `narrowly' ($w_{(t)}$) appeared on the screen one by one. The BOLD signal was recorded by the brain imaging machine after the presentation of these 4 words. Similar arrangements are carried over the other part of the reading process.

We preprocess the fMRI data before training the model to remove noise from the raw data as much as possible.
We compute the default brain activity $\bar{y}$ by selecting the fMRI recording of the default state and averaging these fMRI data. Then we subtract other observed brain activity with the default brain activity $\bar{y}$. The new fMRI data of each time step are used in the experiments.

\subsection{Alignment}

LSTM has two key hidden vector representations to keep the history information and be used in the next time step: (1) a memory vector $c_t$ that summarizes the history of the previous words; and (2) the hidden state vector $h_t$ that is used to predict the probability of the incoming word.


Under the paradigm of the brain-LSTM mapping experiment, the brain activity at the $t$-{th} time step of the story reading process can be viewed as a vector $y_{(t)}$ in the continuous high-dimensional space. Each dimension of $y_{(t)}$ reflects the measured value of the BOLD signal of a certain tiny area (about $3\times3\times3 mm^3$) in the brain, which is also called voxel (VOlumeric piXEL). Mathematically, at the $t$-{th} time step, we use vector $a_{(t)}$ to represent the activations of internal neurons in LSTM. In this paper, $a_{(t)}$ may be memory vector $c_t$ and hidden state vector $h_t$.

To align the activation $a_{(t)}$ of LSTM and brain activity $y_{(t)}$ at the same time step $t$, we  define a function to predict the brain activity $\hat{y}_{(t)}$ from $a_{(t)}$.

In this paper, we use linear function
\begin{align}
{\hat{y}_{(t)}} &=M {a_{(t)}},
\end{align}
where $M$ is the mapping matrix between $a_{(t)}$ and $y_{(t)}$, which is learnt by the least square error.
\begin{align}
M^{\ast} = \argmin_M \sum_{t=1}^{T}{\|{\hat{y}}_{(t)}-y_{(t)}\|}^2.
\end{align}

The matrix $M^{\ast}$ is analytically solved as:
\begin{align}
M^{\ast} = ({A^T}{A})^{-1}{A^T}{Y},
\end{align}
where $A=[a_{(1)},\cdots, a_{(T)}]$ and $Y=[y_{(1)},\cdots, y_{(T)}]$.

Here, we do not train a LSTM neural network to directly predict the brain activities. The reasons lie in two points.

First, the dimension of the fMRI signal varies among different subjects. Therefore, it is not convenient to design a universal neural network architecture for generating outputs of different dimensions.

Second, the goal of this research is not to improve the performance of predicting fMRI signal with LSTM neural network. We just wish to explore the characteristic of the artificial memory vector and the hidden state in the LSTM architecture, as the work on correlating the performance-optimized deep neural network models with the neural activities in the visual cortex \cite{yamins2014performance}.
Therefore, we try to avoid any possible supervision from the fMRI data when training LSTM language model.

\subsection{Evaluation Metric}

Regarding the evaluation metric, we evaluate the model by computing the average cosine distance between the predicted functional brain image and the true observed brain activity at a certain time step of the story reading process. For each activations $a_{(t)}$ of LSTM at the time step $t$ in the test cases, we compute the predicted brain activity $\hat{y_{(t)}}$. Then we calculate the cosine distance between $\hat{y_{(t)}}$ and $y_{(t)}$. Since the cosine distance lies between -1 and 1, we normalise the cosine distance into [0,1] and use it as the accuracy of each test case.
We train the linear map model over about 95\% of the brain imaging data and test the model over the remaining 5\%. We apply 20-folds cross-validation in order to get the average performance of the model.

\begin{figure}[!t]
\pgfplotsset{
tick label style={font=\small},
legend style={font=\scriptsize}
}
\centering
\begin{tikzpicture}
\begin{axis}[
width = 1\linewidth,
height = 0.5\linewidth,
ybar,
bar width = 4pt,
enlargelimits=0.3,
legend style={at={(0.47,0.95)},
anchor=north,legend columns=-1},
ylabel={\small{Similarity}},
ylabel style = {at={(0.05,0.5)}},
xlabel={\small{Test window size}},
symbolic x coords={0,4,400,doc-wise,end},
xtick=data,
]
\addplot[draw,fill=green!50!blue!50!white] coordinates{(doc-wise,0.8313)(400,0.7279)(4,0.4449)};
\addplot[draw,fill=yellow!20!white,pattern=north east lines,pattern color=black!90] coordinates{(doc-wise,0.8651)(400,0.7411)(4,0.4568)};
\addplot[draw,fill=white] coordinates{(doc-wise,0.8490)(400,0.7449)(4,0.4544)};
\addplot[draw,fill=blue!80!black!50!white] coordinates{(doc-wise,0.8906)(400,0.7689)(4,0.4734)};
\addplot[draw,fill=purple!50!white] coordinates{(doc-wise,0.8302)(400,0.7346)(4,0.4643)};
\addplot[draw,fill=brown!60!white] coordinates{(doc-wise,0.8896)(400,0.7611)(4,0.4467)};
\addplot[draw,fill=black!20!white] coordinates{(doc-wise,0.8903)(400,0.7428)(4,0.4597)};
\addplot[draw,fill=blue!20!white,pattern=north west lines,pattern color=blue!90] coordinates{(doc-wise,0.8522)(400,0.7443)(4,0.4577)};

\addplot[red,line legend,sharp plot,
update limits=false]
coordinates {(0,0.4358)(end,0.4358)}
node[above] at (1,0.4358) {\scriptsize{Random (horizontal)}};
\legend{S1,S2,S3,S4,S5,S6,S7,S8}
\end{axis}
\end{tikzpicture}
\caption{{The performance of mapping LSTM memory cell vector to the brain imaging data over different subjects. LSTM model is trained on 8-word window size training data.}}\label{subject-acc}
\end{figure}
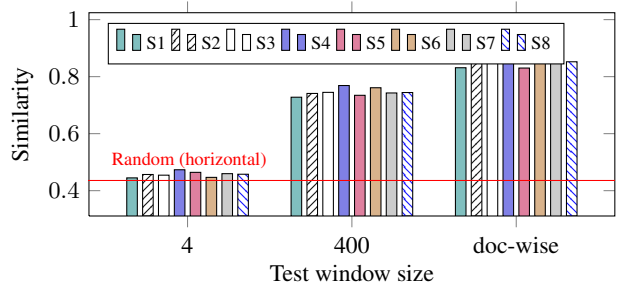

\begin{figure*}[t]
\pgfplotsset{
tick label style={font=\small},
legend style={font=\small}
}


\pgfplotsset{
colormap={hot}{color(0cm)=(blue!80!white); color(1cm)=(pink); color(2cm)=(violet); color(3cm)=(purple)}
}
\begin{center}
\subfloat[{The result of $c_{(t)}$}]{
\begin{tikzpicture}
\begin{axis}[width = 0.41\linewidth,
forget plot style={opacity=0.9},
symbolic x coords={1,4,8,12,16,20,24,28,32,sent-wise},
xtick=data,
symbolic y coords={1,4,8,12,16,20,100,160,200,360,400,600,800,1000,1200,doc-wise},
ytick=data,
xlabel={\small{Train Window Size}},
ylabel={\small{Test Window Size}},
zlabel={\small{Similarity}},
ylabel style = {at={(0.88,0.11)}},
xlabel style = {at={(0.2,-0.02)}},
xlabel style={sloped like x axis},
ylabel style={sloped},
plot box ratio=1 1.8 1.2,
]
\addplot3+[forget plot][surf,scatter,fill=white,mesh/rows=9] coordinates {
(1,1,0.4358) (4,1,0.4525) (8,1,0.4380) (16,1,0.4331)(sent-wise,1,0.4492)
(1,4,0.4537) (4,4,0.4708) (8,4,0.4572) (16,4,0.4537)(sent-wise,4,0.4610)
(1,8,0.4721) (4,8,0.4768)  (8,8,0.4844)(16,8,0.4691) (sent-wise,8,0.4622)
(1,16,0.4766) (4,16,0.4824)(8,16,0.5144)(16,16,0.4841) (sent-wise,16,0.4630)
(1,100,0.5419) (4,100,0.5203)(8,100,0.6032)(16,100,0.5629) (sent-wise,100,0.4764)
(1,200,0.6006) (4,200,0.5414)(8,200,0.6611)(16,200,0.6195) (sent-wise,200,0.5058)
(1,400,0.7508) (4,400,0.7274)(8,400,0.7457)(16,400,0.7530) (sent-wise,400,0.5797)
(1,800,0.8313) (4,800,0.8325)(8,800,0.8204)(16,800,0.8275) (sent-wise,800,0.7015)
(1,doc-wise,0.7870)(4,doc-wise,0.8241)(8,doc-wise,0.8623)(16,doc-wise,0.8173)(sent-wise,doc-wise,0.7570)
};
\end{axis}
\end{tikzpicture}
}
\quad
\subfloat[{The result of $h_{(t)}$}]{
\begin{tikzpicture}
\begin{axis}[width = 0.41\linewidth,
forget plot style={opacity=0.9},
symbolic x coords={1,4,8,12,16,20,24,28,32,sent-wise},
xtick=data,
symbolic y coords={1,4,8,12,16,20,100,160,200,360,400,600,800,1000,1200,doc-wise},
ytick=data,
xlabel={\small{Train Window Size}},
ylabel={\small{Test Window Size}},
zlabel={\small{Similairty}},
ylabel style = {at={(0.88,0.11)}},
xlabel style = {at={(0.2,-0.02)}},
xlabel style={sloped like x axis},
ylabel style={sloped},
plot box ratio=1 1.8 1.2,
]

\addplot3+[forget plot][surf,scatter,fill=white,mesh/rows=9,
] coordinates { 
(1,1,0.4417) (4,1,0.4366) (8,1,0.4370) (16,1,0.4290)(sent-wise,1,0.4449)
(1,4,0.4432) (4,4,0.4303) (8,4,0.4450) (16,4,0.4251)(sent-wise,4,0.4626)
(1,8,0.4572) (4,8,0.4366)  (8,8,0.4482)(16,8,0.4284)(sent-wise,8,0.4599)
(1,16,0.4650) (4,16,0.4366)(8,16,0.4383)(16,16,0.4294)(sent-wise,16,0.4608)
(1,100,0.4690) (4,100,0.4571)(8,100,0.4352)(16,100,0.4294)(sent-wise,100,0.4590)
(1,200,0.4990) (4,200,0.4809)(8,200,0.4314)(16,200,0.4294)(sent-wise,200,0.4568)
(1,400,0.4560) (4,400,0.5500)(8,400,0.4836)(16,400,0.4294)(sent-wise,400,0.4583)
(1,800,0.4953) (4,800,0.5893)(8,800,0.6186)(16,800,0.4294)(sent-wise,800,0.4641)
(1,doc-wise,0.6125)(4,doc-wise,0.6124)(8,doc-wise,0.5437)(16,doc-wise,0.4817)(sent-wise,doc-wise,0.4748)
};
\end{axis}
\end{tikzpicture}
}
\end{center}
\caption{{The similarity between the real brain activities and the ones predicted by (a) the memory vector $c_{(t)}$ and (b) the hidden state vector $h_{(t)}$ of LSTM under different model configurations. The x-axis of each sub-figure represents the window size of the training data. The y-axis of each sub-figure represents the window size of the test data. }}\label{3Dacc}
\end{figure*}
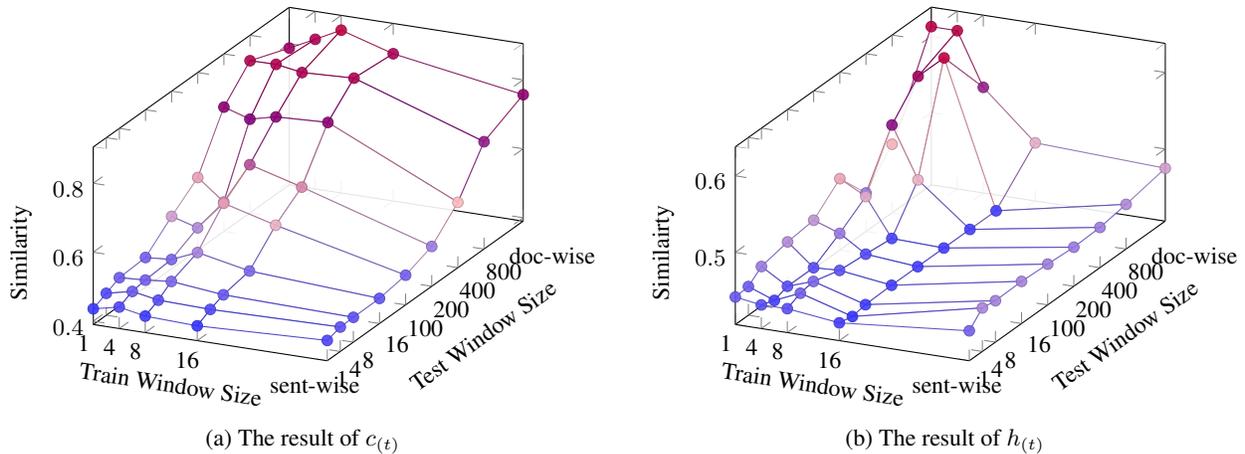

\section{Experiment}

In our experiments, the dimensionality of word embeddings is set to 50, the hidden state size is also set to 50, and the initial learning rate is set to 0.1. The other parameters are initialized by randomly sampling from uniform distribution in [-0.1, 0.1], based on the experience with recurrent neural network.

LSTM is trained according to the procedure of the neural language model \cite{mikolov2010recurrent}, which predicts the incoming word given the history context. Since only one chapter of {\it Harry Porter and the Philosopher's Stone} is involved in the original story reading experiment, the remaining chapters of the book is used as the training data.

Our results report the averaged accuracies over 8 subject under the same experimental conditions. To show the feasibility of this averaging, we show the mean accuracy of LSTM memory vector with three different experimental settings over eight subjects in Figure \ref{acc}. We can see that there is not much between-subject variance. Therefore, each data point in Figure \ref{3Dacc} is computed by averaging the accuracy of 160 test cases (8 subjects with 20 test folds each).

\subsection{Effect of the Long-Term Memory}

LSTM does not explicitly differentiate short-term memory with long-term memory, from a general view of its unit architecture. Therefore, in order to clearly explore how LSTM unit learns to encode the long-term and short-term information and the interaction between the two types of working memory, we deliberately cut the text data with different window size, both for the training corpus and the test stimuli. We set window size as 1, 4, 8, 16 and sentence-length for training data and 1, 4, 8, 16, 100, 200, 400, 800, document-length for test data, as is visualized in Figure \ref{subject-acc}. When training LSTM neural network and generating the vector representation $a_{(t)}$ for every time step $t$, we choose the data of the different window size for LSTM neural network.

The experiment results are presented in Figure \ref{3Dacc}, which suggests that the memory vector of LSTM neural language model generally performs significantly better than the hidden state vector of the neural network in the brain mapping task, given the same hyper-parameter configuration. Besides, the accuracy of the predicted brain image by LSTM memory vector reach about 86\% at the best performance, while the highest accuracy of the predicted brain image by LSTM hidden state vector only reach about 61\%. This supports the cognitive plausibility of the LSTM memory cell architecture.

Regarding the influence of the window size of the training data and the window size of the test data on the model performance, the accuracy increases with large test window size in general. As far as the hidden state vector is concerned, the accuracy also increases with small window size of the training data and the large window size of the test data.

However, we are surprised to find that the memory vector of LSTM architecture achieves the best performance when LSTM model is trained with the text data of exactly the 8-word window size and generate brain activity representation with the word sequence input of the document-wise test window size, as is shown in Figure \ref{3Dacc}. The accuracy sharply decreases when the model generates the representation and predicts fMRI signals only from the previous 4 or 8 words with a limited, small test window size. The accuracy of the memory vector is very low when test window size is small, no matter we decrease or increase the window size of the training data. This indicates that the long-term memory plays an important role in constructing the artificial memory.


\subsection{Effect of the Internal Factor: the Gating Mechanisms}

\begin{figure}[!t]
\centering
\subfloat[{LSTM trained on 8-word window size}]{
\begin{tikzpicture}
\begin{axis}[
width = 0.9\linewidth,height = 0.52\linewidth,
legend style={at={(0.28,0.96)},
anchor=north},
ylabel={\small{Similarity}},ylabel style = {at={(0.04,0.5)}},
symbolic x coords={0W,1,4,8,16,100,200,400,800,doc,AW},
xtick=data,
]
\addplot [color = red!80!white,mark=*]coordinates {(1,0.4380)(4,0.4572)(8,0.4844)(16,0.5144)(100,0.6032)(200,0.6611)(400,0.7457)(800,0.8204)(doc,0.8623)}; 
\addplot [color = black,mark=x]coordinates {(1,0.4405)(4,0.4585)(8,0.4691)(16,0.4720)(100,0.4899)(200,0.5043)(400,0.5795)(800,0.6883)(doc,0.7655)};  
\addplot [color = black,mark=+]coordinates {(1,0.4338)(4,0.4716)(8,0.4871)(16,0.4981)(100,0.5949)(200,0.6388)(400,0.7389)(800,0.8257)(doc,0.8643)};  
\addplot [color = black,mark=triangle]coordinates {(1,0.4447)(4,0.4475)(8,0.4492)(16,0.4506)(100,0.4594)(200,0.4754)(400,0.5153)(800,0.5510)(doc,0.4925)};  
\legend{vanilla {LSTM}, w/o input gate, w/o forget gate, w/o output gate}
\end{axis}
\end{tikzpicture}
}\\

\subfloat[{LSTM trained on 16-word window size}]{
\begin{tikzpicture}
\begin{axis}[
width = 0.9\linewidth,height = 0.52\linewidth,
legend style={at={(0.28,0.96)},
anchor=north},
ylabel={\small{Similarity}},ylabel style = {at={(0.04,0.5)}},
symbolic x coords={0W,1,4,8,16,100,200,400,800,doc,AW},
xtick=data,
]
\addplot [color = red!80!white,mark=*]coordinates {(1,0.4331)(4,0.4537)(8,0.4691)(16,0.4841)(100,0.5629)(200,0.6195)(400,0.7530)(800,0.8275)(doc,0.8173)}; 
\addplot [color = black,mark=x]coordinates {(1,0.4194)(4,0.4069)(8,0.4217)(16,0.4265)(100,0.4950)(200,0.5139)(400,0.5623)(800,0.6323)(doc,0.8111)};  
\addplot [color = black,mark=+]coordinates {(1,0.4394)(4,0.4679)(8,0.4951)(16,0.5340)(100,0.6297)(200,0.7216)(400,0.8196)(800,0.8405)(doc,0.8314)};  
\addplot [color = black,mark=triangle]coordinates {(1,0.4311)(4,0.4461)(8,0.4488)(16,0.4547)(100,0.4754)(200,0.4657)(400,0.4555)(800,0.4500)(doc,0.4545)};  
\legend{vanilla {LSTM}, w/o input gate, w/o forget gate, w/o output gate}
\end{axis}
\end{tikzpicture}
}\\

\subfloat[{LSTM trained on sentence-wise window size}]{
\begin{tikzpicture}
\begin{axis}[
width = 0.9\linewidth,height = 0.52\linewidth,
legend style={at={(0.28,0.96)},
anchor=north},
ylabel={\small{Similarity}},ylabel style = {at={(0.04,0.5)}},
symbolic x coords={0W,1,4,8,16,100,200,400,800,doc,AW},
xtick=data,
]
\addplot [color = red!80!white,mark=*]coordinates {(1,0.4492)(4,0.4610)(8,0.4622)(16,0.4630)(100,0.4764)(200,0.5058)(400,0.5797)(800,0.7015)(doc,0.75)}; 
\addplot [color = black,mark=x]coordinates {(1,0.4303)(4,0.4408)(8,0.4437)(16,0.4527)(100,0.4710)(200,0.5132)(400,0.5898)(800,0.6573)(doc,0.7637)};  
\addplot [color = black,mark=+]coordinates {(1,0.4585)(4,0.4650)(8,0.4821)(16,0.5055)(100,0.5838)(200,0.6451)(400,0.8124)(800,0.8402)(doc,0.8583)};  
\addplot [color = black,mark=triangle]coordinates {(1,0.444787)(4,0.4615)(8,0.4612)(16,0.462)(100,0.4619)(200,0.4620)(400,0.4620)(800,0.4620)(doc,0.4621)};  
\legend{vanilla {LSTM}, w/o input gate, w/o forget gate, w/o output gate}
\end{axis}
\end{tikzpicture}
}
\caption{{Comparative analysis of LSTM model with/without a certain gate. The x-axis of each sub-figure represents the window size of the test data.}}\label{GateAnalysis}
\end{figure}
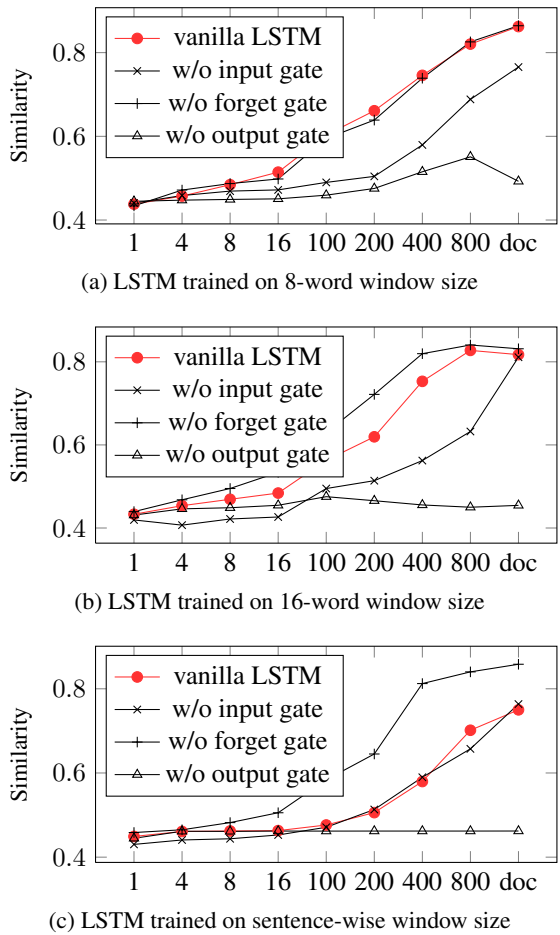

In order to take a further look at the role of each gate in the general LSTM architecture, we ``remove'' the input gates, the forget gates and the output gates respectively by setting their vector as a permanent all-one vector respectively. Then we train the new models with the data of the different window size. The results are presented in Figure \ref{GateAnalysis}. It is obvious that dropping gates brings a fatal influence to the performance on the brain image prediction task.
While it may have negative impact to drop input and output gates, the performances seem to been even improved when dropping the forget gates.

Looking at the brief history of LSTM architecture, we find that the original model in \cite{hochreiter1997long} only has the input gate and the output gate. \cite{gers2000learning} added a forget gate to each LSTM unit, suggesting that it will help improve the model in dealing with continual input stream. It might be the case that story reading, in our experiment, is not a tough prediction task since the size of the input stream is limited to only a part of one novel chapter. In addition, reading should involve the processing of document-wise information, which means setting forget gates to all-one vector should not pose much negative influence to the ability of the model.

It is worth noticing that the performance falls down sharply when the output gates is removed. From a technical perspective, this is probably because that the output gate is close to the hidden state, which means that the output gate receives the back-propagating gradient earlier than the other two gates. The error can not be correctly updated. Therefore, removing the output gate will certainly interfere with the normal back propagation process, leading the model training process towards a wrong direction.

\subsection{Comparison to Other Models}

We compare LSTM model with the vanilla RNN and other heuristic models.
\begin{description}
  \item[Vanilla RNN\_hidden]  We train a vanilla RNN language model and generate a series of representation of the story with the hidden state vector. The experiment configuration of the training and testing data are the same with that of the best LSTM model.
  \item[BoW (tf-idf)] For time step $t$ , we transform the text that the subject has read into a tf-idf (term frequency-inverse document frequency) representation.
  \item[AveEmbedding] We average the word embedding of all the words that have been read at a certain time step $t$ to generate a representation for the brain state. We use the public Turian word embedding dataset \cite{turian2010word}.

\end{description}


Applying the same evaluation metric, we found that the AveEmbed heuristic model performs well, achieving a similarity of 0.81. LSTM memory vector is significantly better than the heuristic method. RNN hidden vectors, however, give out poor performance. A key reason is that RNN only captures short-term information and therefore fails in modelling reading process, which involves strong integration of the long-term information.

\begin{table}[!ht]
\begin{center}
\begin{tabular}{l|cc}
\hline   \multicolumn{1}{c|}{Model} & Cosine Dist. & \multicolumn{1}{c}{Similarity} \\ \hline
Random & -0.128 & 0.436 \\ 
BoW(tf-idf) & 0.184 & 0.592 \\ 
AveEmbedding & 0.634 & 0.817 \\ 
RNN\_hidden & 0.016 & 0.508\\ 
LSTM\_hidden & 0.224 & 0.612 \\ 
LSTM\_memory & \bf{0.724} & \bf{0.862} \\ \hline
\end{tabular}
\end{center}
\caption{{Comparison of different models.}}\label{acc}
\end{table}

\section{Discussion}

We can summary the observations from the experiment results as follows.

\begin{itemize}
  \item LSTM has the ability to encode the semantics of a story by the memory vector, in which the stored information can be used to predict the brain activity with 86\% similairty. Compared to the simple RNN, the overall architecture of LSTM should be more cognitively plausible.
  \item The gating mechanisms are effective for LSTM to filter the valuable information except the forget gates, which is also consistent with the adaptive gating mechanism of working memory system \cite{o2006making}.
  \item The long-term memory can be well kept by LSTM. When we deliberately cut the source of long-term memory (by using small context window size), the prediction accuracy decreases greatly.
\end{itemize}

\begin{figure*}[!t]
  \centering
\subfloat[\small{Real brain activity}]{
\begin{tikzpicture}
    \node[anchor=south west,inner sep=0] (image) at (0,0) {\includegraphics[width=0.33\linewidth]{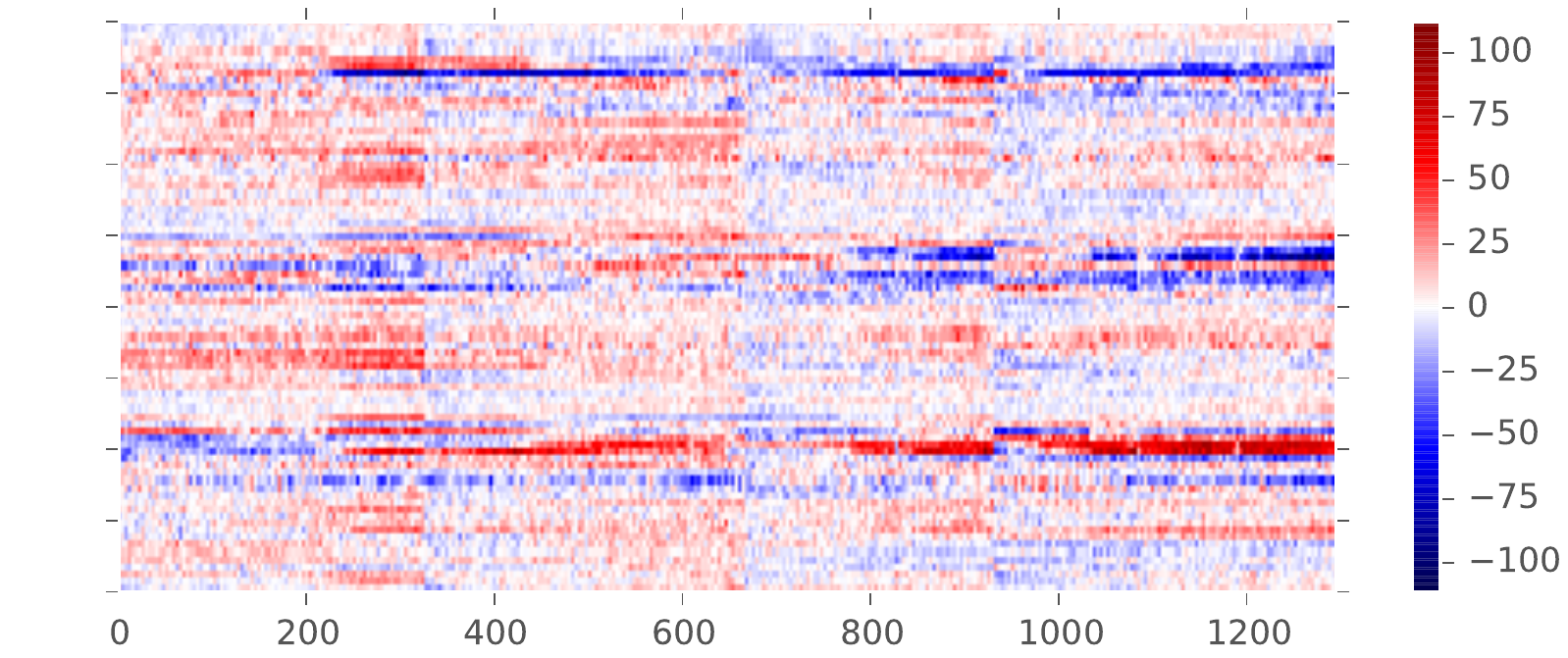}};
    \begin{scope}[x={(image.south east)},y={(image.north west)}]
    \end{scope}
\end{tikzpicture}
}
\subfloat[\small{Predication by LSTM memory cell}]{
\begin{tikzpicture}
    \node[anchor=south west,inner sep=0] (image) at (0,0) {\includegraphics[width=0.33\linewidth]{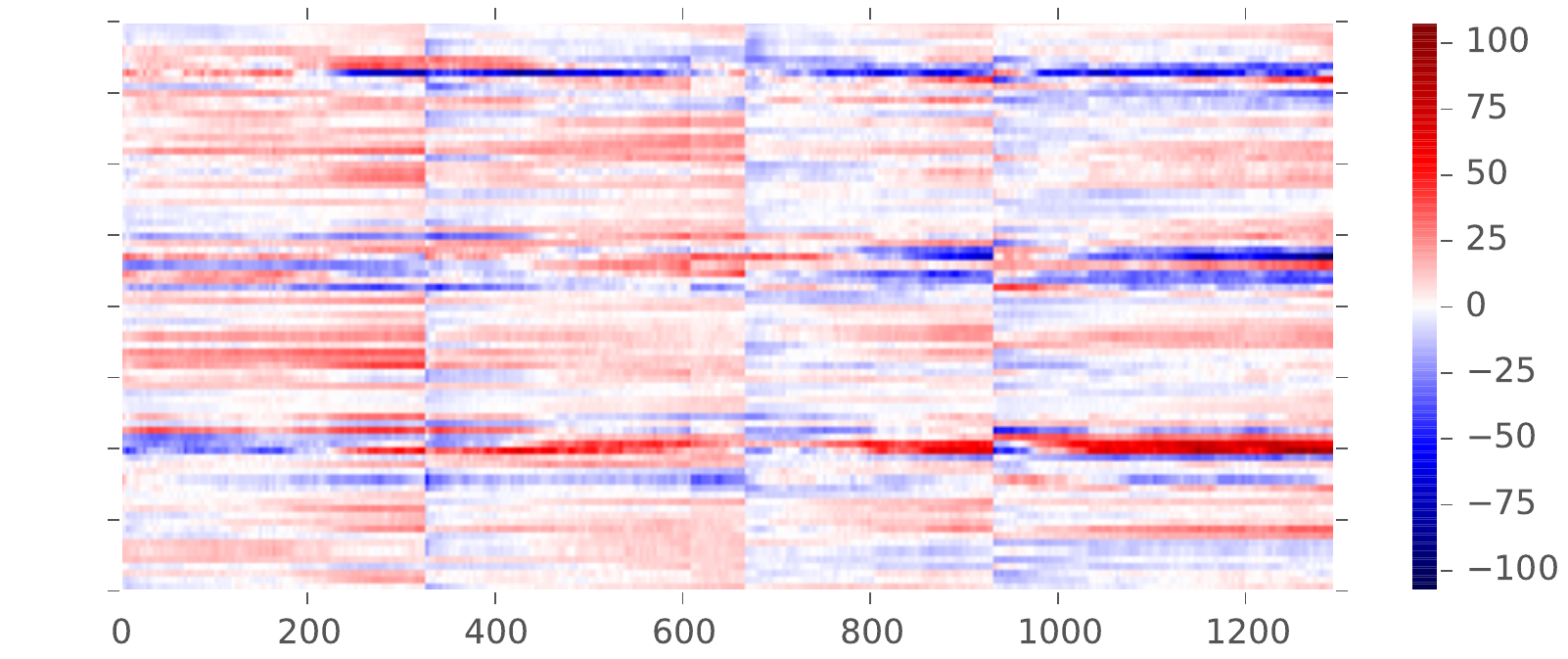}};
    \begin{scope}[x={(image.south east)},y={(image.north west)}]
    \end{scope}
\end{tikzpicture}
}
\subfloat[\small{Predication by LSTM hidden state}]{
\begin{tikzpicture}
    \node[anchor=south west,inner sep=0] (image) at (0,0) {\includegraphics[width=0.33\linewidth]{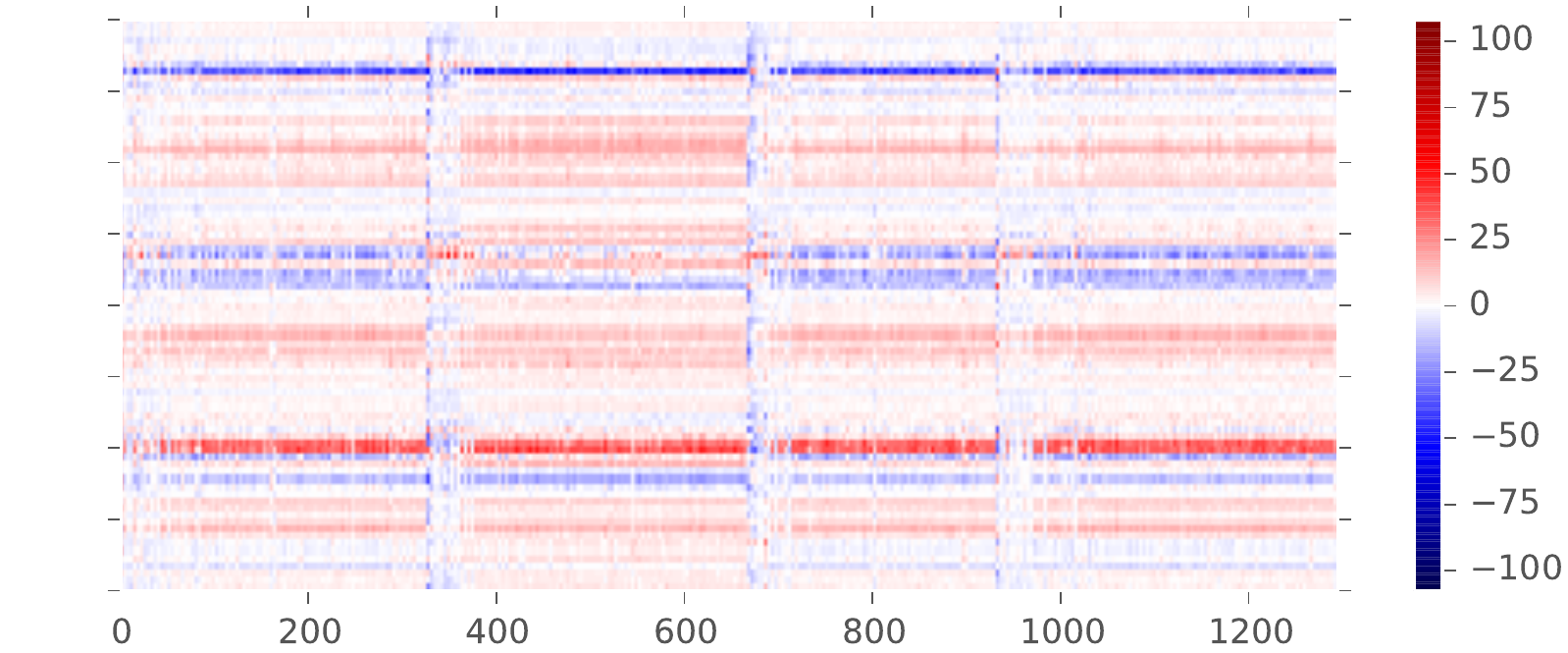}};
    \begin{scope}[x={(image.south east)},y={(image.north west)}]
    \end{scope}
\end{tikzpicture}
}

  \caption{{The real brain activity and the reconstructed brain activities over the subject 1. Four sections of the experiment are concatenated to show the complete time line of the story reading process. The pictures are constructed from (a) true brain data, (b) LSTM memory, and (c) LSTM hidden state.}}\label{brain}
\end{figure*}

%


\begin{figure}[!t]
\tikzset{
every pin/.style={fill=yellow!50!white,rectangle,rounded corners=3pt,font=\tiny},
small dot/.style={fill=yellow!50!white,circle,scale=0.3}
}
\begin{tikzpicture}
    \node[anchor=south west,inner sep=0] (image) at (0,0) {\includegraphics[width=0.95\linewidth]{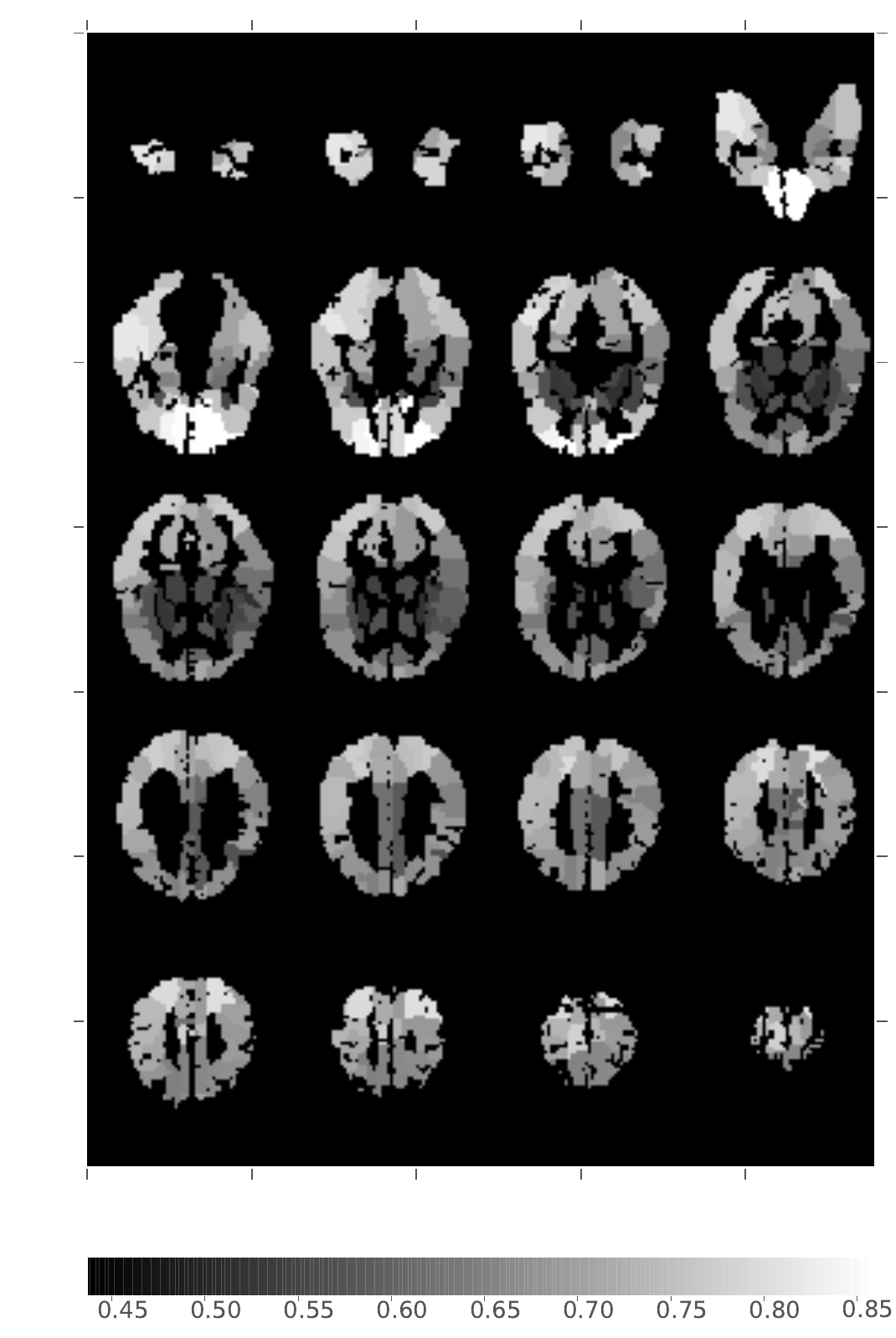}};
    \begin{scope}[x={(image.south east)},y={(image.north west)}]
        \draw[red!50!white,thick] (0.62,0.75) rectangle (0.65,0.78);
        \draw[white] (0.635,0.76) -- +(-0.16in,0.24in)node[anchor=south,font=\scriptsize] { \sffamily Lingual};

        \draw[red!50!white,thick] (0.4,0.66) rectangle (0.42,0.69);
        \draw[white] (0.395,0.675) -- +(-0.15in,-0.0481in)node[anchor=north,font=\scriptsize] { \sffamily Frontal\_Mid\_Orb};
        \draw[red!50!white,thick] (0.59,0.68) rectangle (0.63,0.7);
        \draw[white] (0.61,0.69) -- +(-0.1in,-0.125in)node[anchor=north,font=\scriptsize] { \sffamily Frontal\_Inf\_Orb};

        \draw[red!50!white,thick] (0.58,0.55) rectangle (0.6,0.57);
        \draw[white] (0.59,0.56) -- +(-0.15in,-0.351in)node[anchor=north,font=\scriptsize] { \sffamily Postcentral};

        \draw[red!50!white,thick] (0.38,0.76) rectangle (0.41,0.80);
        \draw[white] (0.395,0.775) -- +(-0.14in,0.19in)node[anchor=south,font=\scriptsize] { \sffamily Fusiform};

        \draw[red!50!white,thick] (0.37,0.87) rectangle (0.42,0.89);
        \draw[white] (0.4,0.88) -- +(0.1in,0.18in)node[anchor=south,font=\scriptsize]{\sffamily Temproal\_Pole\_Mid};

        \draw[red!50!white,thick] (0.8,0.9) rectangle (0.84,0.93);
        \draw[white] (0.82,0.91) -- +(0in,0.15in)node[anchor=south,font=\scriptsize] { \sffamily Temporal\_Inf};

        \draw[red!50!white,thick] (0.86,0.84) rectangle (0.91,0.86);
        \draw[white] (0.9,0.86) -- +(-0.3in,-0.08in)node[anchor=north,font=\scriptsize] { \sffamily Frontal\_Sup\_Orb};
        \draw[red!50!white,thick] (0.38,0.24) rectangle (0.41,0.27);
        \draw[white] (0.4,0.25) -- +(0in,0.15in)node[white,anchor=south,font=\scriptsize]{\sffamily Parietal\_Sup};
        \draw (0.13,0.96)node[white]{\small \bfseries L};
        \draw (0.95,0.96)node[white]{\small \bfseries R};
    \end{scope}
\end{tikzpicture}
\caption{{The correlation between the predicted and the real neural activity in a certain anatomical region for Subject 1. The brain is displayed in the transverse view. }}\label{roi-corr}
\end{figure}
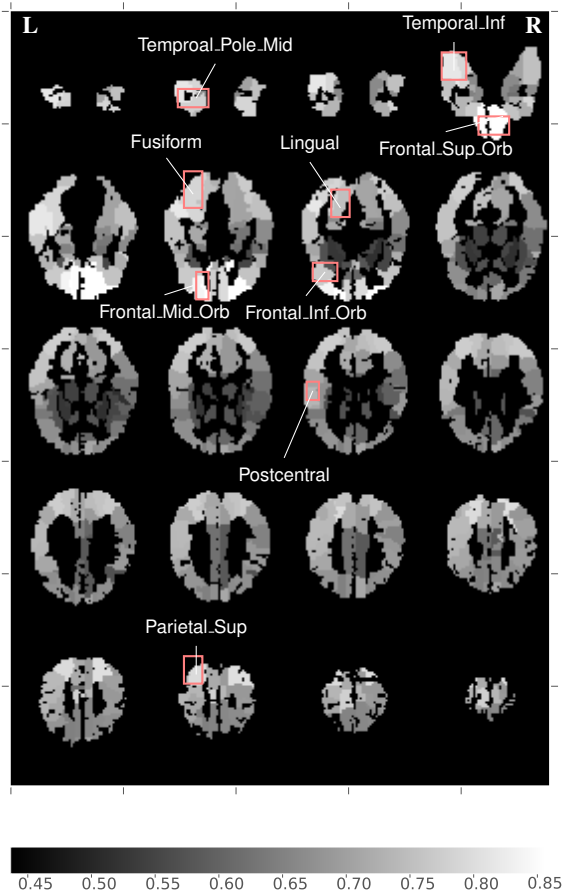

\subsection{Visualization Analysis}

In addition to the quantitative analysis above, we visualize part of the brain state dynamics of subject 1 in Figure \ref{brain}, including the true signal sequence at 80 randomly-selected voxels and two signal sequences reconstructed by LSTM memory vector and LSTM hidden state vector. The x-axis of each sub-figure represents the time steps of the reading process (the fixation periods between every two runs are removed).
The colour indicates the activation level of a certain brain region at a certain time step.

We found that the brain dynamics reconstructed by the memory vector of LSTM is more like a smoothed version of the real brain activity.
The brain dynamics reconstructed by the hidden state vector of LSTM are largely deviated from the real brain activity, although LSTM hidden state reconstructs a few features of the real brain signals.



\subsection{Connection with Cognitive Neuroscience}



To explore whether the predictability of the neural activity varies among different brain regions for LSTM memory vector, we compute the correlation of the predicted and the real brain activities by measuring the Pearson correlation coefficient for each voxel within each subject. Then we compute the averaged correlation for a specific anatomical region defined by AAL (Automated Anatomical Labeling) atlas and visualize the correlation strength for subject 1 in Figure \ref{roi-corr}.

We notice some interesting phenomena that might reflect the association of the LSTM memory vector with the predictability of brain regions involved in language processing and semantic working memory.

\cite{gabrieli1998role} indicates that prefrontal cortex is associated with semantic working memory. 
\cite{price2000anatomy} summarizes that frontal superior gyrus is associated with the processing of word meaning and that temporal pole area is associated with sentence reading. Our analysis also reflects a strong correlation between the reconstructed brain activity from LSTM memory vector and the observed brain activity of the prefrontal cortex and temporal cortex, especially the inferior and anterior part of the gyrus.

We also found that the reconstructed brain activity of Lingual gyrus and Fusiform (Visual Word Form Area ) are highly correlated with the real observed activities. Previous neuroscience research \cite{mechelli2000differential,mccandliss2003visual} has reported that these brain regions play an important role in word recognition.
Similar patterns have also been found for other subjects.


\section{Conclusion}

In this paper, we explore LSTM architecture with the sequential brain signal of story reading.
Experiment results suggest a correlation between the LSTM memory cell and the cognitive process of story reading.
In the future work, we will continue to investigate the effectiveness of different LSTM variants by relating the representation generated by the models with neural dynamics. We would also try to design a more reasonable artificial memory architecture for a better approximation to the working memory system and language cognition. Besides, we will investigate some non-linear mapping function between the artificial and brain memories.


\section*{Acknowledgments}

We would like to thank the anonymous reviewers for their valuable comments. This work was partially funded by National Natural Science Foundation of China (No. 61532011, 61473092, and 61472088), the National High Technology Research and Development Program of China (No. 2015AA015408).


\end{document}